\documentclass[letterpaper, 10 pt, conference]{ieeeconf}  

\IEEEoverridecommandlockouts                              

\usepackage{graphics} 
\usepackage{epsfig} 
\usepackage{times} 
\usepackage{amsmath} 
\usepackage{amssymb}  
\usepackage{amsfonts}
\usepackage{import}
\usepackage{subcaption}
\usepackage[dvipsnames]{xcolor}
\usepackage{array}
\usepackage{booktabs}
\usepackage{multirow}
\usepackage[margin=2cm]{geometry}
\usepackage{adjustbox}
\usepackage{soul}
\usepackage{url}
\usepackage{hyperref}

\DeclareMathOperator*{\argmax}{argmax}
\DeclareMathOperator*{\argmin}{argmin}

\title{\LARGE \bf
NVINS: Robust Visual Inertial Navigation Fused with NeRF-augmented Camera Pose Regressor and Uncertainty Quantification
}

\author{Juyeop Han$^{1}$, Lukas Lao Beyer$^{1}$, Guilherme V. Cavalheiro$^{1}$, and Sertac Karaman$^{1}$
\thanks{$^{1}$The authors are with Laboratory for Information and Decision Systems (LIDS), Massachusetts Institute of Technology, Cambridge, MA 02139, USA
        {\tt\small \{juyeop, llb, guivenca ,sertac\}@mit.edu}}%
}

\begin{document}

\maketitle
\thispagestyle{empty}
\pagestyle{empty}

\newcommand{\first}[1]{\textcolor{blue}{\textbf{#1}}}
\newcommand{\second}[1]{\textcolor{red}{\textbf{#1}}}
\begin{abstract}

In recent years, Neural Radiance Fields (NeRF) have emerged as a powerful tool for 3D reconstruction and novel view synthesis. However, the computational cost of NeRF rendering and degradation in quality due to the presence of artifacts pose significant challenges for its application in real-time and robust robotic tasks, especially on embedded systems. This paper introduces a novel framework that integrates NeRF-derived localization information with Visual-Inertial Odometry (VIO) to provide a robust solution for real-time robotic navigation. By training an absolute pose regression network with augmented image data rendered from a NeRF and quantifying its uncertainty, our approach effectively counters positional drift and enhances system reliability. We also establish a mathematically sound foundation for combining visual inertial navigation with camera localization neural networks, considering uncertainty under a Bayesian framework. Experimental validation in a photorealistic simulation environment demonstrates significant improvements in accuracy compared to a conventional VIO approach.

\end{abstract}


\section{Introduction}

A Neural Radiance Field (NeRF) is a recent type of machine learning method that can learn continuous representations of a scene's color and density properties given a set of training images, which can allow for the synthesis of high quality novel views \cite{NeRF20ECCV}.
Their impressive initial results, despite limitations, have garnered the attention of researchers. For robotic applications, NeRFs and similar methods provide a conceptually flexible way to fuse camera measurements into a representation that does not suffer from the curse of dimensionality and synthesizes useful information, such as an environment's geometry.

Preliminary research has started exploring the applicability of these representations in perception tasks such as localization \cite{Lin2021inerf, Moreau22LENS}, navigation \cite{maggio23ICRA, Adamkiewicz22RAL,  katragadda2023nerfvins} and SLAM \cite{Wiesmann2023LocNDF, Sucar2021iMAP, Zhu22niceslam, Rosinol2023Nerfslam}.
General NeRF improvements and variations were also suggested in order to address some of their main issues, such as their significant computational requirements \cite{FridovichKeil2022plenoxels, Kerbl3DGaussians, mueller2022instant}, quality degradation under multi-scale data \cite{Barron2021mipnerf} and failures under a few-shot setting \cite{yang2023freenerf}.
Nevertheless, many challenges remain for robotics applications due to limited computational resources in embedded systems and the occurrence of artifacts and reconstruction errors, all the more common in less controlled environments.

In this paper, we propose a new framework to extract information from an imperfect NeRF representation of an environment and integrate it into a visual-inertial odometry (VIO) solution using a rigorous mathematical foundation, while maintaining computational feasibility for onboard systems.
To circumvent the computational demands of the NeRF's rendering process, we distill its localization information into a Convolutional Neural Network (CNN) trained offline on generating pairs of poses and images.
Similarly to \cite{Moreau22LENS}, the CNN estimates the absolute pose associated with an image, thus providing evidence against position drift, akin to a GPS sensor.
In addition, we incorporate uncertainty quantification and outlier rejection strategies to allow for tight integration with a VIO pipeline.
We evaluate methods such as Monte Carlo (MC) dropout \cite{Gal2016MCdropout} and deep ensembles \cite{Lakshminarayanan2017Ensemble} as options for uncertainty quantification and evaluate the resulting navigation performance for several trajectories simulated in a photo-realistic environment \cite{Guerra2019Flightgoggles}.

The main contributions of this research are summarized as below:
\begin{itemize}
	\item{We propose a real-time and loop closure-free VIO framework leveraging information extracted from an imperfectly trained NeRF that can be feasibly run on embedded hardware.}
	\item{We mathematically formulate the integration between VIO and the uncertain poses estimated by a neural network as \textit{Maximum a Posteriori} (MAP) optimization in a Bayesian setting.}
	\item{We analyze the accuracy and efficiency improvements of the proposed framework compared to baselines.}
\end{itemize}

\begin{figure}[tb!]
	\centering
	\begin{subfigure}{.45\textwidth} \label{subfig:offline}
		\includegraphics[width=\linewidth]{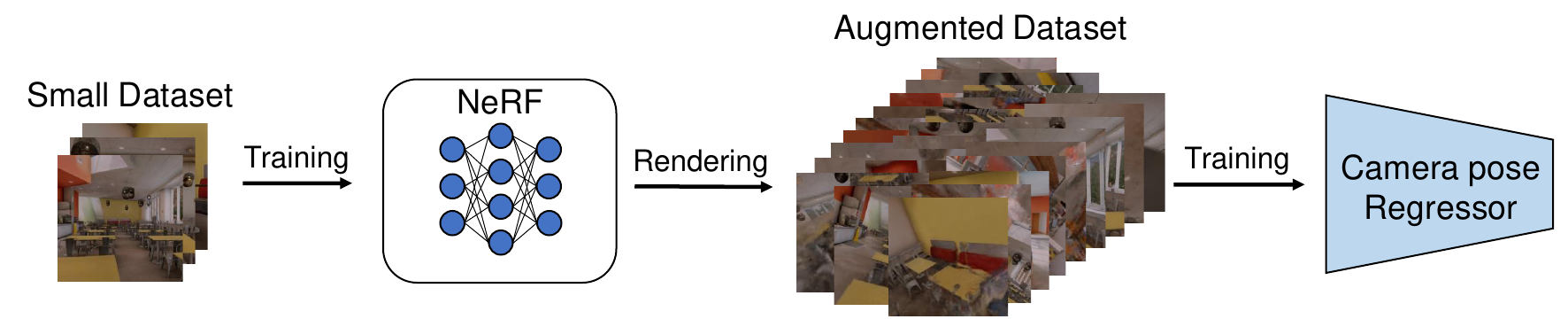}
		\caption{offline}
	\end{subfigure}
	\\
	\begin{subfigure}{.40\textwidth} \label{subfig:online}
		\includegraphics[width=\linewidth]{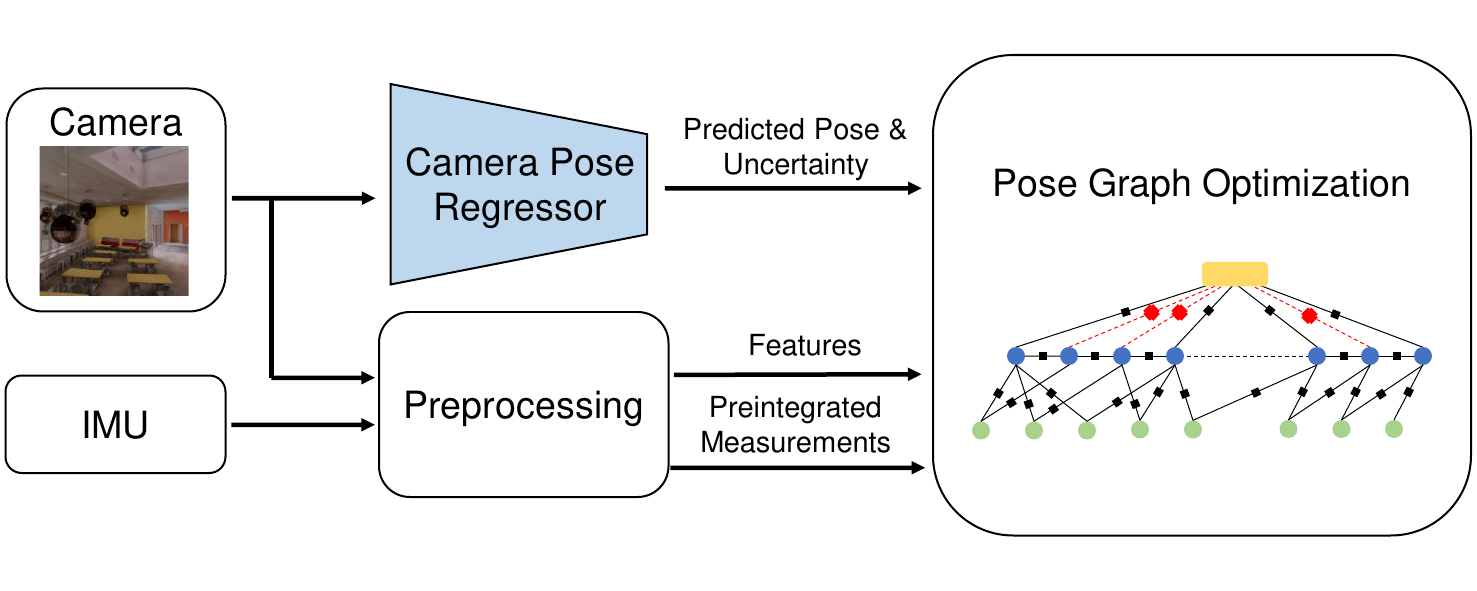}
		\caption{online}
	\end{subfigure}
	\caption{Outline of the proposed VIO framework, NVINS: (a) In the offline phase, a NeRF is trained using a dataset gathered in the target environment. Novel views synthesized using this NeRF are used to train the camera pose regressor. (b) A pose predicted by the trained camera pose regressor, along with its associated uncertainty, is integrated with other sensor measurements via pose graph optimization.} \label{fig:outline}
	\vspace{-0.2in}
\end{figure}

\section{Related Work}

\textbf{Robot Perception with Neural Scene Representation.}
Neural scene representations like NeRF have significantly advanced robotic perception capabilities, particularly in camera-only localization tasks. Techniques such as iNeRF \cite{Lin2021inerf} optimize camera poses by minimizing pixel residuals between real and NeRF-rendered images. Loc-NeRF \cite{maggio23ICRA} and LENS \cite{Moreau22LENS} enhance localization through particle filters and augmented data, respectively, while frameworks like that of Adamkiewicz \textit{et al.} \cite{Adamkiewicz22RAL} integrate dynamics with NeRF for vision-only planning.

In dense SLAM, iMAP \cite{Sucar2021iMAP}, NICE-SLAM \cite{Zhu22niceslam} employ MLPs for pose estimation and mapping, albeit requiring depth data from RGB-D cameras.
NeRF-SLAM \cite{Rosinol2023Nerfslam} applies dense monocular SLAM to achieve faster and more precise scene representation compared to other methods.
NeRF-VINS \cite{katragadda2023nerfvins} tightly couples pre-trained NeRF with VIO by comparing features from the real image and the image rendered by NeRF and performs in real-time on an onboard machine.
However, the match between features of real and rendered images remains questionable in cases of low-resolution rendered images \cite{ReviewNeRF} or artifacts, often resulting from inadequate training, such as using sparse images.

\textbf{CNN-based Camera Pose Regression and Uncertainty Quantification.}
CNNs are widely used for camera pose estimation from images, with PoseNet~\cite{Kendall2015Posenet, Kendall2016ICRA, Kendall2017CVPR} being the pioneering CNN-based regressor. Research has explored structural enhancements like encoder-decoder~\cite{Melekhov2017ICCV} and LSTM architectures~\cite{Walch2017ICCV}, and training improvements through prediction of relative camera poses~\cite{Brahmbhatt2018Mapnet}.

Uncertainty in camera pose regression is captured through the covariance matrix of the pose, distinguishing between \textit{aleatoric} and \textit{epistemic uncertainty}~\cite{Kendall2017Nips}. Aleatoric uncertainty refers to the irreducible uncertainty associated with the noise inherent in the measurements. Epistemic uncertainty is the reducible uncertainty stemming from a lack of knowledge beyond the training dataset.
Kendall \textit{et al.}~\cite{Kendall2016ICRA} quantify epistemic uncertainty using MC dropout to estimate errors caused by out-of-distribution inputs. To combine uncertainty estimated by the pose regressor with additional sensor measurements, conventional methods such as the Kalman filter \cite{Moreau2022Coordinet} or visual odometry (VO) \cite{peretroukhin2020hydranet} have been used.
Although epistemic and aleatoric uncertainty have been simultaneously estimated in prior work~\cite{peretroukhin2020hydranet}, we are not aware of any system that thoroughly utilizes this type of 6-DoF uncertainty estimate in a pose graph optimization framework.

\begin{figure}[tb!]
	\centering
	\begin{subfigure}{.20\textwidth}
		\includegraphics[width=\linewidth]{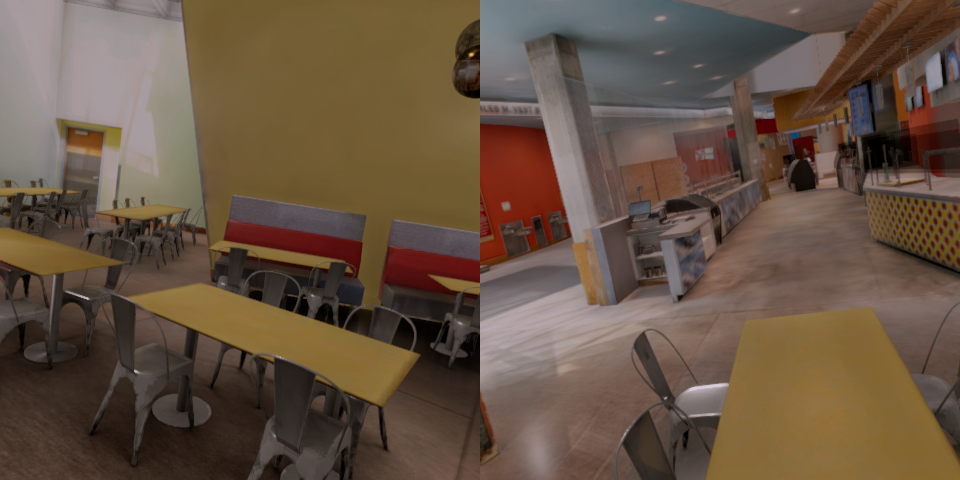}
		\caption{FlightGoggles}
	\end{subfigure}
	\begin{subfigure}{.20\textwidth}
		\includegraphics[width=\linewidth]{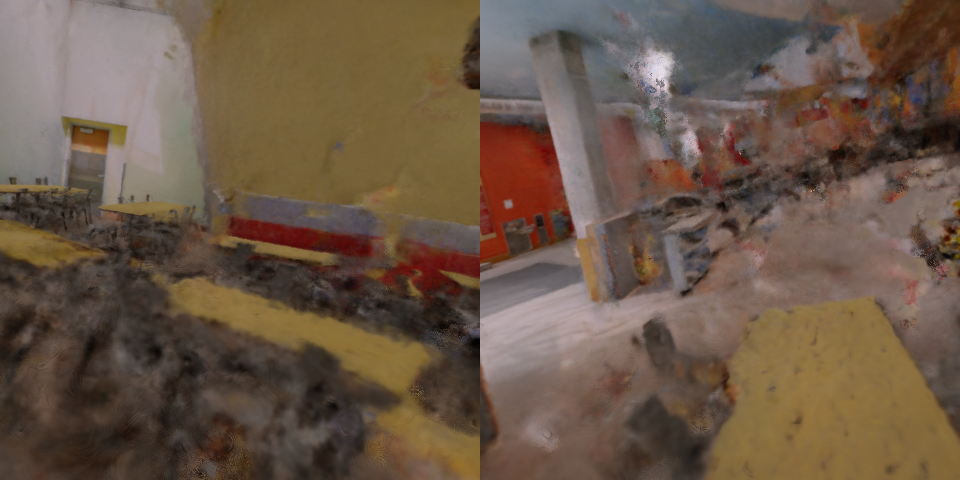}
		\caption{NeRF}
	\end{subfigure}
	\caption{(a) Images captured in the Flightgoggles simulation environment \cite{Guerra2019Flightgoggles}. (b) Images rendered through nerfstudio \cite{nerfstudio}. The images exhibit significant differences from the originals, primarily due to artifacts caused by the sparse density of training data.} \label{fig:sim_and_render}
	\vspace{-0.2in}
\end{figure}

\section{Preliminaries} \label{sec:preliminaries}
In this section, the formulation of two types of Bayesian Neural Networks (BNN) -- deep ensemble and MC dropout -- is presented.
Note that deep ensemble can also be interpreted from a Bayesian perspective~\cite{Wilson2020Neurips}, even though deep ensemble was initially described as a non-Bayesian technique~\cite{Lakshminarayanan2017Ensemble}.

Under the BNN framework, the neural network's weights, $\mathbf{w}$, are not considered fixed values. Instead, they initially follow a prior probabilistic distribution, $P(\mathbf{w})$. For a given training dataset $\mathcal{D} = \{ \mathcal{X}, \mathcal{Y} \}$, where $\mathcal{X}$ represents the input features and $\mathcal{Y}$ the target variable, these weights conform to a posterior distribution, $P(\mathbf{w} \vert \mathcal{D})$.

\textbf{Deep Ensemble \cite{Lakshminarayanan2017Ensemble, Wilson2020Neurips}}: In deep ensemble, a finite set of weights, $\mathcal{W}_p = \{ \mathbf{w}_i^p \}$, is sampled from the prior distribution of weights, $P(\mathbf{w})$.
Each sampled weight, $\mathbf{w}_i^p$, is optimized to find a maximum a posteriori (MAP) estimate $ \mathbf{w}_i^* = \argmax_{\mathbf{w}}P(\mathbf{w} \vert \mathcal{D})= \argmin_{\mathbf{w}}(-\log P( \mathcal{Y} \vert \mathbf{w}, \mathcal{X})P(\mathbf{w}) )$ where $P( \mathcal{Y} \vert \mathbf{w}, \mathcal{X})$ is a likelihood of the label data $\mathcal{Y}$ given the input data $\mathcal{X}$ and weights $\mathbf{w}$.
Then, the posterior distribution $P(\mathbf{w} \vert \mathcal{D})$ is approximated as
\begin{equation}
q(\mathbf{w}) \approx \frac{1}{\vert \mathcal{W}_p \vert}\sum_{i=1}^{\vert \mathcal{W}_p \vert} \delta(\mathbf{w} = \mathbf{w}_i^*)
\end{equation}
where $\delta(\cdot)$ is a Dirac delta function.

The negative log-likelihood of the posterior is used to construct the loss function
\begin{equation}
\label{eq:total_loss}
\mathcal{L}_{total}(\mathbf{w}) = \mathcal{L}(\mathbf{w}) + \mathcal{L}_{reg}(\mathbf{w}),
\end{equation}
where the two components, $\mathcal{L}(\mathbf{w})$ and $\mathcal{L}_{reg}(\mathbf{w})$, correspond to the negative log-likelihood of  $P(\mathcal{Y} \vert \mathbf{w}, \mathcal{X})$ and $P(\mathbf{w})$, respectively.

\textbf{Monte Carlo (MC) dropout \cite{Gal2016MCdropout}}: MC dropout is a variational inference method which approximates the true posterior distribution over weights~$P(\mathbf{w} \vert \mathcal{D})$ as~$q_{\theta^*}(\mathbf{w})$, a neural network with model parameters $\theta^*$.
At test time, weights are sampled from $q_{\theta^*}(\mathbf{w})$, i.e., $\hat{\mathbf{w}} = \{ \hat{\mathbf{W}}_i \}_{i=1}^L \sim q_{\theta^*}(\mathbf{w})$.
The weight matrix of $i$-th layer, $\hat{\mathbf{W}}_i$, is dropped out with a probability $1 - p_i$.
That is, the weights at each layer $i$ follow the Bernoulli distribution:
\begin{align}
\mathbf{b}_j^{i} &\sim \text{Bernoulli}(p_i) \quad \forall j \in \{1, \cdots, K_{i-1}\}, \\
\hat{\mathbf{W}}_i &=  \Theta_i \cdot \text{diag}(\mathbf{b}^{i}).
\end{align}
Here, $K_i$ is the number of neurons at the layer $i$.
To apply MC dropout at layer $i$, a vector of independent Bernoulli samples $\mathbf{b}^{i}$ is drawn.
If $j$-th element of $\mathbf{b}^{i}$ is zero (i.e., $\mathbf{b}_j^{i} = 0$), the corresponding weight is dropped out.

We can find the optimal parameter $\theta^*$ minimizing the KL divergence $\text{KL}(q_{\theta}(\mathbf{w}) \Vert P(\mathbf{w} \vert \mathcal{D}))$, which directly corresponds to minimize $\mathcal{L}_{total}(\theta)$ with replacing weights $\mathbf{w}$ to the set of weight matrices, $\theta = \{ \Theta_i \}_{i=1}^L$ \cite{Kendall2017Nips}.

\section{Training NeRF and Camera Pose Regressor with Uncertainty} \label{sec:train}

In this section, NeRF is initially trained with a small dataset $\mathcal{D}_O = \{\mathcal{I}_O, \mathcal{T}_O\}$ to generate much larger dataset, $\mathcal{D}_A = \{\mathcal{I}_A, \mathcal{T}_A \}$, by rendering images corresponding to sampled poses using NeRF.
Here, the dataset $\mathcal{D}_{\{O, A\}} = \{\mathcal{I}_{\{O, A\}}, \mathcal{T}_{\{O, A\}} \}$ consists of a set of images, $\mathcal{I}_{\{O, A\}}$, and their corresponding poses of camera, $\mathcal{T}_{\{O, A\}}$.
Subsequently, the posterior of the camera pose $T_n \in \text{SE(3)}$ given the dataset $\mathcal{D}_A$ and the input image $I_n$, $P(T_n \vert \mathcal{D}_A, I_n)$, is derived in the form of a Bayesian neural network (BNN) while quantifying its uncertainty.
The output of the neural network and its uncertainty will be employed to compute the optimal state of VIO in Section \ref{sec:map}.

\subsection{Training NeRF and Image Rendering}
\label{subsec:NeRF}
We use a NeRF reconstruction as a form of data augmentation, generating a large dataset $\mathcal{D}_A = \{\mathcal{I}_A, \mathcal{T}_A\}$ from a relatively small dataset $\mathcal{D}_O = \{\mathcal{I}_O, \mathcal{T}_O\}$. Even if the original poses $\mathcal{T}_O$ are not accurate, they can be refined using bundle adjustment techniques such as COLMAP~\cite{schoenberger2016sfm}.
A neural network $[\mathbf{c}, \sigma] = f_{\phi}(\mathbf{x}, \mathbf{d})$ parameterized by $\phi$ is trained with $\mathcal{D}_O$. The network takes two inputs: $\mathbf{x} \in \mathbb{R}^3$ and $\mathbf{d} \in \mathbb{R}^2$ representing the position and the direction of the ray for rendering, respectively.
The outputs, $\mathbf{c}$ and $\sigma$, represent the view-dependent color and view-independent volume density of the sampled point of the ray.
An intensity of one pixel is obtained through rendering a ray.
The loss function of the neural network $f_{\phi}(\mathbf{x}, \mathbf{d})$ is designed to minimize the gap between the pixel intensity of the camera image and the estimated pixel intensity obtained through rendering process.
For more information about NeRF, we refer to \cite{NeRF20ECCV}.

Once the NeRF reconstruction $f_\phi$ has been obtained, the augmented dataset $\mathcal{D}_A$ is constructed by rendering camera images $\mathcal{I}$ for corresponding camera poses~$\mathcal{T}$.
We obtain the set of poses~$\mathcal{T}$ by sampling within a predefined area of interest.
Note that, while we choose to use NeRF for data augmentation, other novel view synthesis techniques such as \cite{FridovichKeil2022plenoxels} and \cite{Kerbl3DGaussians} can also be employed.

\subsection{Camera Pose Regression with Uncertainty Quantification}
\label{subsec:poseregrssor}

In order to estimate a camera pose~$T_n$ given a query camera image~$I_n$, we define the neural network $f_{\mathbf{w}}: \mathbb{R}^{W \times H \times 3} \rightarrow \mathbb{R}^{12}$ trained on the augmented dataset $\mathcal{D}_A = \{\mathcal{I}_A, \mathcal{T}_A \}$, where W and H are the width and height dimension of the image, and $\mathbf{w}$ is the weight of the neural network:
\begin{equation} \label{eq:network}
[\hat{\mathbf{t}}_n^T, (\boldsymbol{\sigma}_{n,p}^{a})^T]^T = [f_{\mathbf{w}}^t(I_n), f_{\mathbf{w}}^{\sigma}(I_n)]^T = f_{\mathbf{w}}(I_n)
\end{equation}
$I_n \in \mathbb{R}^{W \times H \times 3}$ is the input of the function $f_{\mathbf{w}}$.
$\hat{\mathbf{t}}_n$ and $\boldsymbol{\sigma}_{n,p}^{a}$ are the vectorized representation of the pose and aleatoric uncertainty of the position, respectively.
$f_{\mathbf{w}}$ utilizes a CNN to process high-dimensional image data. Notably, pixel coordinates are added as extra channels before each convolutional layer to improve CNN performance on coordinate transformation-related tasks \cite{Liu2018CoordConv}.
The vectorized pose representation, $\hat{\mathbf{t}}_n = [\hat{\mathbf{p}}_n^T,  \hat{\mathbf{r}}_n^T]^T \in \mathbb{R}^9$, consists of a 3D position and a 6D representation of rotation. This choice avoid the high errors associated with discontinuous representations such as axis-angle or quaternion \cite{Zhou2019CVPR}.
We approximate the covariance of the position $\mathbf{p}_n$ as a diagonal matrix, i.e. $S_{n,p}^{a} = \text{diag}(\boldsymbol{\sigma}_{n,p}^{a})$.

We assume that the neural network (\ref{eq:network}), $f_{\mathbf{w}}$,  models the distribution of estimated pose as a combination of Gaussian distribution of position and isotropic Langevin distribution of rotation given the weight $\mathbf{w}$ and query image $I_n$ as an input:
\begin{equation} \label{eq:gaussian}
	P(\mathbf{t}_n \vert I_n, \mathbf{w})
= \frac{\exp\big(
		-\frac{1}{2}\Vert \mathbf{p}_n - \hat{\mathbf{p}}_n \Vert_{S^a_n}^2
		+ \kappa \text{tr}(\hat{R}_n^TR_n) \big)}{c(\kappa)\sqrt{(2\pi)^3\det{(S^a_{n,p})}}}
\end{equation}
with a rotation matrix $R_n \in \text{SO(3)}$ and a normalizing factor, $c(\kappa)$.
$S_n^{a}$ and $\kappa^{-1}$ are interpreted {as an \textit{aleatoric uncertainty}, an irreducible uncertainty caused by a random noise.
Note that \textit{epistemic uncertainty}, which will be represented later, has the same dimension as the aleatoric uncertainty to reflect the pose distribution (\ref{eq:gaussian}).
In this research, we treat the inverse of the rotational aleatoric uncertainty $\kappa$ as a known constant due to the absence of an analytical form for $c(\kappa)$, which complicates the modeling of $c(\kappa)$  \cite{Chiuso2008Infosys}.

We leverage the BNN to model an \textit{epistemic uncertainty}, which arises when the model encounters inputs that deviate from the training dataset $\mathcal{D}_A$.
As shown in Fig. 2, rendered images $\mathcal{I_A}$ cannot be perfectly identical to the original. When the model takes the images in the real environment, the gap between the rendered images and the original images is modeled as epistemic uncertainty.
Unlike conventional neural networks, in a BNN, the weight $\mathbf{w}$ does not assume a fixed value but follows a posterior weight distribution $P(\mathbf{w} \vert \mathcal{D}_A)$ given the dataset $\mathcal{D}_A$, as discussed in Sec. \ref{sec:preliminaries}.
To approximate the posterior weight distribution, we adopt deep ensemble \cite{Lakshminarayanan2017Ensemble} and MC dropout \cite{Gal2016MCdropout} due to their simplicity and scalability compared to other approaches \cite{BNNReview}.

The loss function for the weight $\mathbf{w}$ is formulated as shown in (\ref{eq:total_loss}). Assuming independence between each data pair $(I_n, \mathbf{t}_n)$, the loss function $\mathcal{L}(\mathbf{w})$ corresponding to the negative log-likelihood of $P(\mathcal{T}_A \vert \mathcal{I}_A, \mathbf{w})$ is represented as
\begin{equation} \label{eq:nn_loss}
\begin{split}
\mathcal{L}(\mathbf{w}) = &\frac{1}{N}\sum_{n=1}^N
\frac{1}{2} 
\Big( \kappa\Vert R_n - \hat{R}_n \Vert^2_F + \\
& \Vert \mathbf{p}_n - \hat{\mathbf{p}}_n \Vert^2_{S_{n,p}^a} + \text{logdet}(S_{n,p}^a) \Big).
\end{split}
\end{equation}
For numerical stability, the covariance output of the network $\boldsymbol{\sigma}_n^a$ is replaced with the log vector of the covariance output, $\mathbf{s}_{n,p}^a$, such that $\sigma_{n,p,m}^a = \exp(s_{n,p,m}^a)$.
Both the estimated mean, $\hat{\mathbf{t}}_n$, and log of variance, $\mathbf{s}_{n,p}^a$, are parameterized by $\mathbf{w}$. Additionally, the regularization loss, $\mathcal{L}_{reg}(\mathbf{w})$, assumes the prior distribution $P(\mathbf{w})$ is Gaussian with zero mean, formulated as $\mathcal{L}_{reg}(\mathbf{w}) = \lambda \Vert \mathbf{w} \Vert_2^2$ with a positive constant $\lambda$.

A distribution of a camera pose, $\mathbf{t}_n$, given the input image, $I_n$, and training dataset $\mathcal{D}_A$ is derived by marginalizing over the weight of BNN:
\begin{equation} \label{eq:marginalization}
P(\mathbf{t}_n \vert \mathcal{D}_A, I_n) = \int_{\mathbf{w}} P(\mathbf{t}_n \vert I_n, \mathbf{w}) P(\mathbf{w} \vert \mathcal{D}_A)d\mathbf{w}
\end{equation}

For practical implementation, a weight $\mathbf{w}_i$ is sampled from the distribution $q(\mathbf{w})$, approximating $P(\mathbf{w} \vert \mathcal{D}_A)$ through either deep ensemble or MC dropout. This results in a set of weights $\mathcal{W} = \{\mathbf{w}_i\}$.
The pose distribution $P(\mathbf{t}_n \vert \mathcal{D}_A, I_n)$ is parameterized as the average and covariance of the outputs from neural networks with the sampled weights, represented as $f_{\mathbf{w}_i}(I_n) = [(\mathbf{t}_n^i)^T \; (\boldsymbol{\sigma}_{n,p}^{a,i})^T]^T, \ \forall \mathbf{w}_i \in \mathcal{W}$.
The averages of the position and the orientation are estimated as:
\begin{equation}
	\begin{gathered}
	\bar{\mathbf{p}}_n = \frac{1}{\vert \mathcal{W} \vert}\sum_{w_i \in \mathcal{W}}\mathbf{p}_n^i, \;
	\bar{R} = \argmin_{R \in \text{SO(3)}} \sum_{i=1}^{\vert \mathcal{W} \vert} \Vert R_n^i - R \Vert^2_F
	\end{gathered}
\end{equation}

The covariance, $\hat{\Sigma}_{n, \{p, r\} } = \hat{\Sigma}_{n, \{p, r\}}^a + \hat{\Sigma}_{n, \{p, r\}}^e$}, of the pose distribution (\ref{eq:marginalization}) is estimated as the sum of the aleatoric uncertainty, $\hat{\Sigma}_{n, \{p, r\}}^a$,  and the epistemic uncertainty, $\hat{\Sigma}_{n, \{p, r\}}^e$ with $\{ p, \, r \}$ representing position and rotation \cite{Kendall2017Nips}.
Assuming the distribution of $\Vert R_n - \bar{R}_n \Vert_F$ can be approximated by a zero-mean Gaussian distribution, the uncertainties can be represented as follows:
\begin{equation}
\begin{gathered}
\hat{\Sigma}_{n,p}^a
= \frac{\sum_{w_i \in \mathcal{W}}S_n^{a,i}}{\vert \mathcal{W} \vert - 1},
\hat{\Sigma}_{n,p}^e = \frac{\sum_{w_i \in \mathcal{W}}\tilde{\mathbf{p}}_n^i(\tilde{\mathbf{p}}_n^i)^T}{\vert \mathcal{W} \vert - 1}\\
\hat{\Sigma}_{n,r}^a = \kappa^{-1},
\hat{\Sigma}_{n,r}^e = \frac{\sum_{w_i \in \mathcal{W}}\Vert R_n^i - \bar{R}_n \Vert_F^2}{\vert \mathcal{W} \vert - 1}
\end{gathered}
\end{equation}
with $\tilde{\mathbf{p}}_n^i = \mathbf{p}_n^i - \bar{\mathbf{p}}_n$. The estimated average, $\bar{T}_n \in \text{SE(3)}$ corresponding $\bar{\mathbf{p}}_n$ and $\bar{R}$, and covariance, $\hat{\Sigma}_n$, of the pose will be utilized into the VIO framework to compute the residual function (\ref{eq:residuals}).

\section{VIO Integration with Camera Pose Regressor} \label{sec:map}

In this section, we describe how the BNN's absolute pose estimate is incorporated into a visual-inertial odometry (VIO) framework.
We view this problem as maximum a posteriori (MAP) inference on a factor graph.
In particular, we aim to acquire the optimal estimate of the state $\mathcal{X} = \{\mathcal{X}_B, \; \mathbf{P} \}$:
\begin{equation}\label{eq:MAP_0}
\mathcal{X}^* = \argmax_{\mathcal{X}}P(\mathcal{X} \vert \mathcal{D}_A, \mathcal{Z}),
\end{equation}
through factor graph optimization, where
\begin{align}
  \mathcal{X}_B &= \{\mathbf{x}_B^{(1)}, \cdots, \mathbf{x}_B^{(N)} \}, &
  \mathbf{P} &= [\mathbf{p}_1 \cdots \mathbf{p}_M]^T, \\
  \mathcal{D}_A &= \{\mathcal{I}_A, \mathcal{T}_A \}, \text{and} &
  \mathcal{Z} &= \{\mathcal{Z}_B, \mathcal{Z}_C, \mathcal{I}_o \}.
\end{align}
Here, $N$ and $M$ denote the number of poses in the discretized trajectory estimate and the number of observed landmarks, respectively.
Each pose $\mathbf{x}_B^{(n)} \in \text{SE}(3)$ describes the world-to-body transform at time $t_n$.
The collection of observed landmarks is denoted by~$\mathbf{P} \in \mathbb{R}^{M \times 3}$.
The augmented training dataset introduced in Section~\ref{sec:train} and the sensor inputs (composed of IMU measurements $\mathcal{Z}_B$, tracked features extracted by a VIO frontend $\mathcal{Z}_C$, and camera images $\mathcal{I}_o$) are denoted by $\mathcal{D}_A$ and $\mathcal{Z}$, respectively.
We model the IMU according to \cite{Forster2017TRO}, and have omitted details such as velocity and bias walk estimation of IMU for notational simplicity.

Moreover, we selectively exclude the neural network output identified as outliers from the optimization, ensuring a more robust MAP estimation using the uncertainty of the neural network.

\subsection{Factor Graph Optimization} \label{subsec:MAP}

Assuming zero-mean Gaussian noise for all measurements and that all measurements are independent of each other except the relation between the offline data $\mathcal{D}_A$ and camera images $\mathcal{I}_o$, the MAP estimate (\ref{eq:MAP_0}) is given by
\begin{equation} \label{eq:MAP_1}
\begin{split}
\mathcal{X}^*
	= \argmax_{\mathcal{X}}& P(\mathcal{X} \vert \mathcal{D}_A, \mathcal{I}_o)P(\mathcal{Z}_B, \mathcal{Z}_C \vert \mathcal{X}, \mathcal{D}_A, \mathcal{I}_o) \\
 	= \argmax_{\mathcal{X}}& P(\mathcal{X} \vert \mathcal{D}_A, \mathcal{I}_o) P(\mathcal{Z}_B \vert \mathcal{X})P(\mathcal{Z}_C \vert \mathcal{X}) \\
 	= \argmax_{\mathcal{X}}& \Big\{ \prod_{n \in \mathcal{N}}P(T_n \vert \mathcal{D}_A, I_n) \\
 	&\prod_{b \in \mathcal{B}}P(\mathbf{z}_{b_{k+1}}^{b_{k}}  \vert \mathcal{X})
 	\prod_{(i,j) \in \mathcal{C}}P(\mathbf{z}^{j}_i \vert \mathcal{X}) \Big\}.
\end{split}
\end{equation}
The posterior camera pose $P(T_n \vert \mathcal{D}_A, I_n)$ is given by the output of the BNN~(\ref{eq:marginalization}).
$P(\mathbf{z}_{b_{k+1}}^{b_{k}}  \vert \mathcal{X})$ and $P(\mathbf{z}^{j}_i \vert \mathcal{X})$ are likelihoods of IMU and image feature measurements, where $\mathbf{z}^{j}_i \in \mathbb{R}^2$ refers to the $i$-th feature captured by the $j$-th camera frame.
Note that we use IMU preintegration~\cite{Forster2017TRO}, so that $\mathbf{z}_{b_{k+1}}^{b_{k}}$ corresponds to the integrated IMU measurement in the time interval $[t_k, t_{k+1})$.

The nonlinear optimization problem~(\ref{eq:MAP_1}) is solved by minimizing the sum of Mahalanobis distances:
\begin{equation} \label{eq:nonlinear_opt}
\begin{split}
&\mathcal{X}^* = \argmin_{\mathcal{X}}\Big\{
\sum_{n \in \mathcal{N}} \Vert \mathbf{r}_\mathcal{N}(I_n, \mathcal{X}) \Vert_{\hat{\Sigma}_n}^2 \\
&+ \sum_{k \in \mathcal{B}} \Vert \mathbf{r}_\mathcal{B}(\mathbf{z}^{b_k}_{b_{k+1}}, \mathcal{X}) \Vert_{\Sigma^{b_k}_{b_{k+1}}}^2
+ \sum_{(i, j) \in \mathcal{C}} \Vert \mathbf{r}_\mathcal{C}(\mathbf{z}^{j}_i, \mathcal{X}) \Vert_{\Sigma^{j}_l}^2
\Big\}
\end{split}
\end{equation}

The residual $\mathbf{r}_{\mathcal{N}}(I_n, \mathcal{X})$ is defined as the combination of the Euclidean position error and the Frobenius norm of the rotational error, $\bar{T}_n$, obtained from the BNN, and the actual camera pose, $T_n$:
\begin{equation}
\label{eq:residuals}
\mathbf{r}_{\mathcal{N}}(I_n, \mathcal{X}) = [(\mathbf{p}_n - \bar{\mathbf{p}}_n)^T, \, \Vert R_n - \bar{R}_n \Vert_F]^T.
\end{equation}
For convenience of implementation, $\Vert R_n - \bar{R}_n \Vert_F$ in (\ref{eq:residuals}) is replaced with $\sqrt{2}\Vert \text{Log}(\bar{R}_n^TR_n) \Vert_2$, since
\begin{equation}
\Vert \text{Log}(\bar{R}_n^TR_n) \Vert_2 = \epsilon, \, \Vert R_n - \bar{R}_n \Vert_F = 2\sqrt{2}\sin \frac{\epsilon}{2} \approx \sqrt{2}\epsilon,
\end{equation}
where $\epsilon$ is the angle between the two rotations, and $\text{Log}(\cdot)$ refers to the logarithm map on SO(3). $\mathbf{r}_{\mathcal{B}}(\mathbf{z}_{b_k}^{b{k+1}}, \mathcal{X})$ and $\mathbf{r}_{\mathcal{C}}(\mathbf{z}_i^j, \mathcal{X})$ represent the residuals based on IMU measurements \cite{Forster2017TRO} and the camera feature measurements \cite{BA2000}, respectively.

\subsection{Uncertainty-based Outlier Rejection} \label{subsec:outlier}
As shown in Fig. \ref{fig:sim_and_render}, areas with little or no coverage in the original, unaugmented dataset~$\mathcal{D}_{\text{NeRF}}$ cannot be reconstructed accurately.
We find that the pose and uncertainty~$\hat{\Sigma}_{n, (\cdot)}$ predicted by our BNN can be unreliable in such cases.
Thus, we propose two simple and effective rejection heuristics to detect and discard pose regressor outputs that are deemed unreliable:
\begin{equation} \label{eq:threshold}
\text{tr}({\hat{\Sigma}_{n,p}}) > \tau_1
\quad \text{or} \quad
\Vert \mathbf{p}_n^* - \mathbf{\hat{p}}_n \Vert_{\hat{\Sigma}_{n,p}} > \tau_2.
\end{equation}
First, we exclude predictions for which the magnitude of the trace of the positional uncertainty exceeds a certain threshold~$\tau_1$.
Additionally, we compute the Mahalanobis distance between the most recent pose estimated by the VIO system and the current pose predicted by the pose regressor, under the covariance as predicted by the pose regressor.
When this distance exceeds a predefined threshold~$\tau_2$, the BNN output is discarded.

\section{Experiments}

Experiments are conducted to validate (a) improved pose accuracy achieved through increased data with NeRF and uncertainty quantification, (b) consistent pose estimation without the need for loop closure, and (c) computational efficiency of the proposed VIO framework. All experiments are conducted within the FlightGoggles simulation~\cite{Guerra2019Flightgoggles}.

We compare the accuracy of the camera pose regressors trained with and without uncertainty prediction. We also assess the computational efficiency of the designed pose regressors on both a desktop and a Jetson AGX Xavier onboard machine (Sec. \ref{subsec:exp_neuralnet}). The accuracy of our proposed VIO framework is then measured under different settings regarding uncertainties (Sec. \ref{subsec:exp_vio}).

We collected 1,297 posed 480 px$\times$480 px RGB-D images within a 15m$\times$24m$\times$4m region of the ``Stata Center'' model available in the Flightgoggles simulation~\cite{Guerra2019Flightgoggles}.
This dataset is used to train the \textit{depth-nerfacto} model~\cite{nerfstudio}, which incorporates the multiresolution hash encoding from Instant-NGP~\cite{mueller2022instant} for improved performance and can make use of depth information.

Upon completing the training of the NeRF model, 50,000 pairs of poses and images were generated by the trained NeRF model and utilized to train the camera pose regressors.
The orientations of the sampled poses are constrained in order to avoid extreme attitudes exceeding pitch or roll angles of approximately 30 degrees.
The camera pose regressor architecture includes 4 CoordConv layers \cite{Liu2018CoordConv} followed by 2 (resp. 3) fully connected layers for the pose (resp. uncertainty) output.
We constructed a relatively shallow network compared to other pose regressors \cite{Kendall2015Posenet, Brahmbhatt2018Mapnet, Moreau2022WACV} to enable real-time onboard inference.
Training of the pose regressor networks was conducted using the Adam optimizer with a learning rate of $10^{-3}$ for 100 epochs.
The weighting factor~$\kappa$ (see~(\ref{eq:nn_loss})) was set to 1.
The weight decay factor for all networks is set to $\lambda = 10^{-5}$.
These parameters were selected through trial and error to maximize the performance of the networks.

\begin{table}[tbp]
	\centering
	\caption{Inference speeds of camera pose regressors on desktop and Jetson AGX Xavier (ms)}
	\centering
	\begin{tabular}{c || c | c }
		\toprule
		\textbf{Networks} & \textbf{Desktop} & \textbf{Jetson AGX Xavier}  \\
		\midrule
		\textbf{no-nerf} & 1.638 $\pm$ 0.122 & 8.378 $\pm$ 1.951  \\
		\textbf{vanilla} & 1.603 $\pm$ 0.063 & 8.299 $\pm$ 0.866 \\
		\textbf{dropout} & 13.989 $\pm$ 0.691 & 41.27 $\pm$ 1.949 \\
		\textbf{ensemble} & 8.255 $\pm$ 0.166 & 29.91 $\pm$ 1.247 \\
		\bottomrule
	\end{tabular} \label{tab:computation}
\end{table}

\begin{figure}[tb!]
	\centering
	\includegraphics[width=\columnwidth]{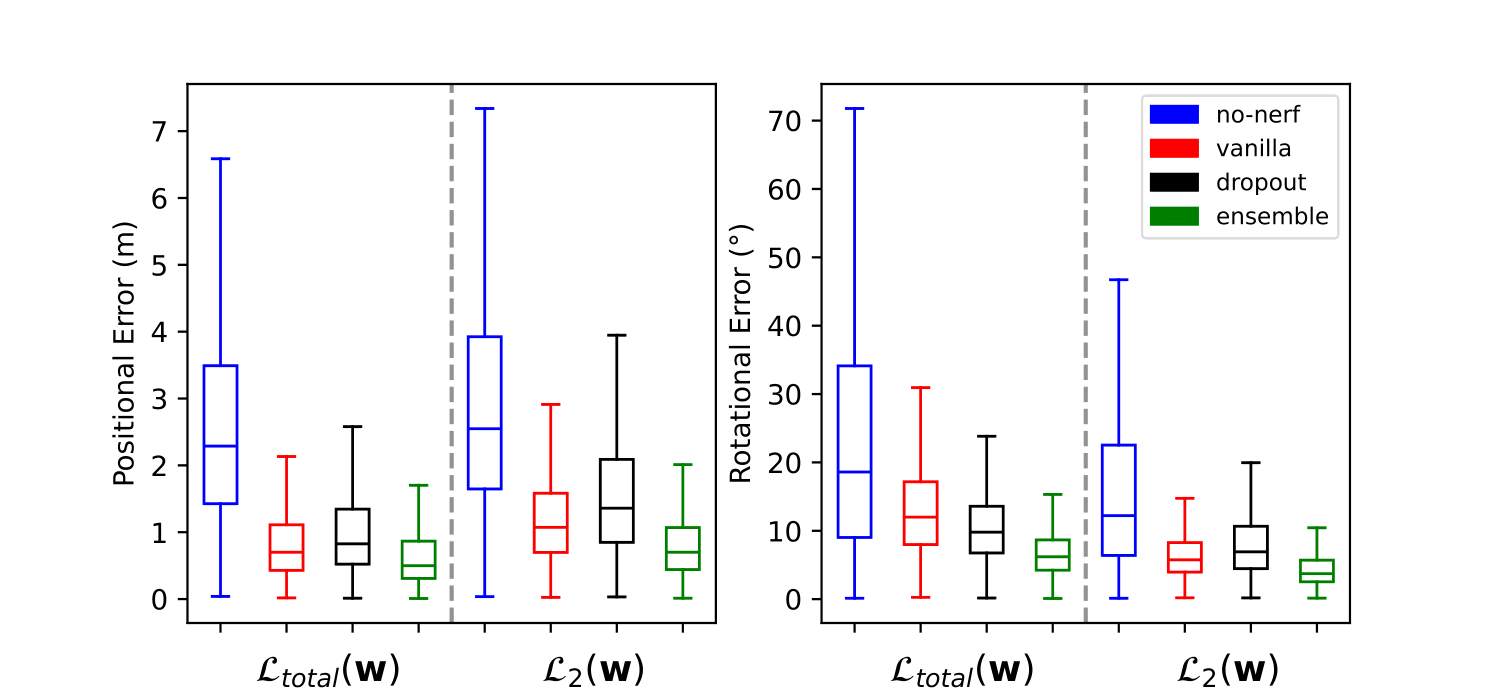}
	\caption{Positional and rotational errors of various camera pose regressors trained with different loss functions across entire trajectories (m/degree)} \label{fig:NN_accuracy}
	\vspace{-0.2in}
\end{figure}

\begin{figure*}[htb!]
	\centering
	\includegraphics[width=2\columnwidth]{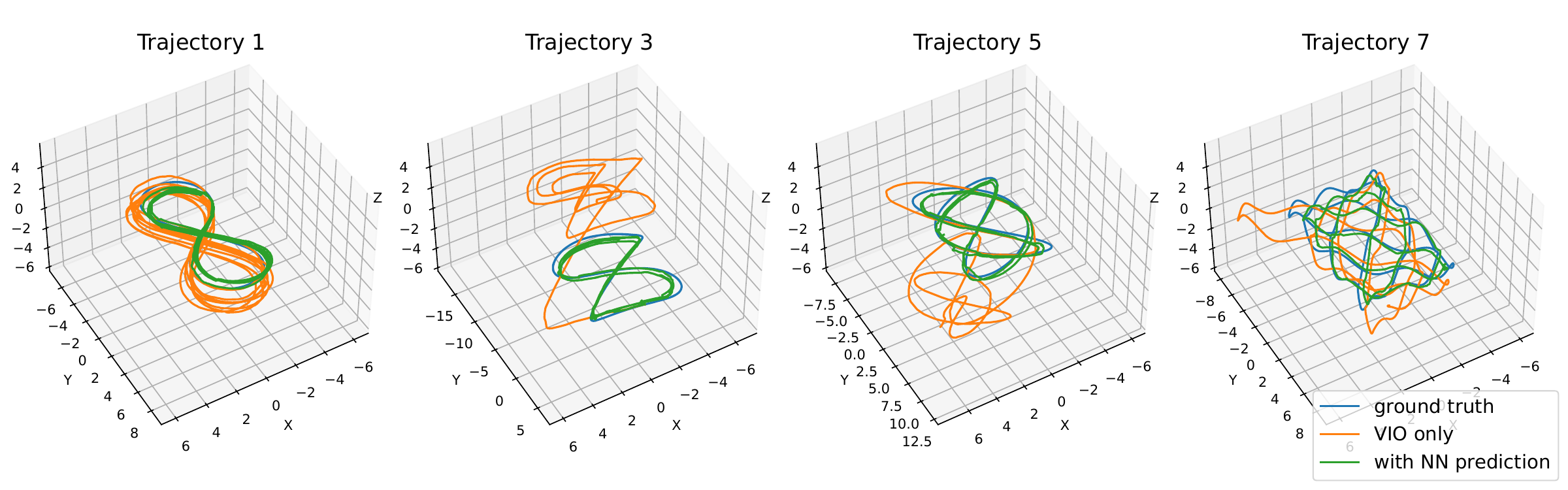}
	\caption{Comparison of ground truth trajectories, `Trajectory 1, 3, 5, and 7', (blue) with estimations from the VIO framework without the camera pose regressor (orange), and with the ensemble camera pose regressor augmented by uncertainty quantification and outlier rejection (green)} \label{fig:traj}
	\vspace{-0.2in}
\end{figure*}

\subsection{Camera Pose Regressor with Uncertainty} \label{subsec:exp_neuralnet}

\textbf{Setup.} In this subsection, we aim to validate (a) the improved performance through the increased number of training data aided by NeRF and adoption MC dropout and deep ensemble, (b) the effect of aleatoric uncertainty output $\boldsymbol{\sigma}^a_{n,p}$, and (c) computation speed of the camera pose regressor across different computation environment.

We trained pose regressors with two distinct loss functions: $\mathcal{L}_{total}(\mathbf{w})$ as defined in (\ref{eq:total_loss}) and $\mathcal{L}_{2}(\mathbf{w})$ which is identical to $\mathcal{L}_{total}(\mathbf{w})$ except that the positional covariance $S_{n,p}^a$ of $\mathcal{L}_{2}(\mathbf{w})$ is considered constant.
For each loss function, four camera pose regressors were trained and named as follows: \textbf{no-nerf}, \textbf{vanilla}, \textbf{dropout}, and \textbf{ensemble}. `No-nerf' and `vanilla' are neural networks trained with the dataset used to train the NeRF model and the dataset rendered by NeRF, respectively. `Dropout' and `ensemble' are neural networks designed to quantify the uncertainties of the outputs, utilizing MC dropout and deep ensemble, respectively. For the dropout implementation, the dropout rate was uniformly set at $p = 0.05$ for all layers. `Ensemble' has 8 independent neural networks trained, each with their parameters initialized according to a Gaussian distribution with a standard deviation of $\alpha = 0.02$. To ensure a fair comparison, `dropout' had 8 forward passes for inference.

After training, the pose regressors predict the camera pose attached to a quadrotor across seven different trajectories for approximately two minutes within the simulation environment. One trajectory follows the Lemniscate of Bernoulli (\textbf{Trajectory 1}), while the others are Lissajous curves, each with distinct parameters (\textbf{Trajectory 2 - 7}).

\textbf{Results.}
Firstly, the inference speeds of the camera pose regressors, tested on `Trajectory 1', are compared on a desktop equipped with Intel Core i9-7920X CPU at 2.90GHz and NVIDIA TITAN V 12GB GPU, and on an NVIDIA Jetson AGX Xavier, as shown in Table. \ref{tab:computation}. Although the `dropout' and `ensemble' demonstrate inference speeds a few times slower than those of `no-nerf' and `vanilla', they are still sufficiently fast for VIO integration in real-time considering that the slowest inference speed, observed with `dropout' on the Jetson AGX Javier, achieves about 24Hz.

Fig. \ref{fig:NN_accuracy} displays the errors for the trained pose regressors having loss functions $\mathcal{L}_{2}(\mathbf{w})$ and $\mathcal{L}_{total}(\mathbf{w})$, assessed across the whole testing trajectories. It is evident that generating training data via NeRF significantly increases the accuracy of the pose regressors. Furthermore, it is shown that pose regressors trained with $\mathcal{L}_{total}(\mathbf{w})$ exhibit better positional accuracy than those with $\mathcal{L}_{2}(\mathbf{w})$. On the other hand, pose regressors trained with $\mathcal{L}_{total}(\mathbf{w})$ demonstrate worse rotational accuracy. It is attributed to the neural network's uncertainty output, automatically weighing losses for position and rotation in $\mathcal{L}_{total}(\mathbf{w})$. Additionally, `ensemble' consistently outperforms the other methods, contrary to `dropout' which underperforms relative to 'vanilla' in terms of positional accuracy. The performance of `dropout' might be improved by applying dropout to fewer layers, as suggested in \cite{Kendall2016ICRA}.

\subsection{Integration VIO with Camera Pose Regressor} \label{subsec:exp_vio}

\textbf{Setup.}
Each trained pose regressor provided the VIO framework with a sequence of estimated camera poses and their uncertainties, $\{\bar{T}_n, \hat{\Sigma}_n\}_{n \in \mathcal{N}}$ used to compute the sum of residual norms of the residual functions (\ref{eq:nonlinear_opt}, \ref{eq:residuals}). We employed 'dropout' and `ensemble' regressors trained with $\mathcal{L}_{total}(\mathbf{w})$.

To evaluate the effectiveness of incorporating the camera pose regressor into the VIO framework—specifically, to examine the impact of including uncertainty predictions on noise and deviation from the training dataset—we conducted tests across the seven trajectories described in the preceding subsection. Additionally, we established a baseline by testing the VIO framework without the camera pose regressor, referred to as \textbf{only VIO}. These tests were conducted under eight different scenarios, based on three criteria, named as \textbf{(option 1)} + \textbf{(option 2)} + \textbf{(option 3)}: \textit{(option 1)} type of pose regressors (`dropout' or `ensemble'), \textit{(option 2)} the application of predicted uncertainty $\hat{\Sigma}_n$ within the VIO framework (`constant' or 'estimated'), and \textit{(option 3)} the application of outlier rejection ('no rejection' and 'rejection'). Note that loop closure is not applied in any of the cases.

When `constant' and 'rejection' are concurrently applied, only the Mahalanobis norm-based rejection is implemented, as the trace remains unchanged.
In the `constant' scenario, a fixed covariance, $\text{diag}([0.2 ; 0.2 ; 0.2 ; 1.0]^T)$, is utilized instead of $\hat{\Sigma}_n$.
The thresholds for outlier rejection~(\ref{eq:threshold}) are set at $\tau_1 = 1.0$ and $\tau_2 = 0.4$.

For the VIO pipeline, features are detected using GFTT~\cite{gftt}  and tracked using the Lucas-Kanade algorithm~\cite{lucas1981iterative}.
The backend, using GTSAM~\cite{gtsam}, optimizes factor graphs at 17 Hz, combining features, 240 Hz IMU data, and pose regressor estimates. Pose regressor estimates are incorporated at every alternate keyframe, except when rejected based on specific heuristics~(\ref{eq:threshold}).

\begin{figure}[tb!]
	\centering
	\includegraphics[width=\columnwidth]{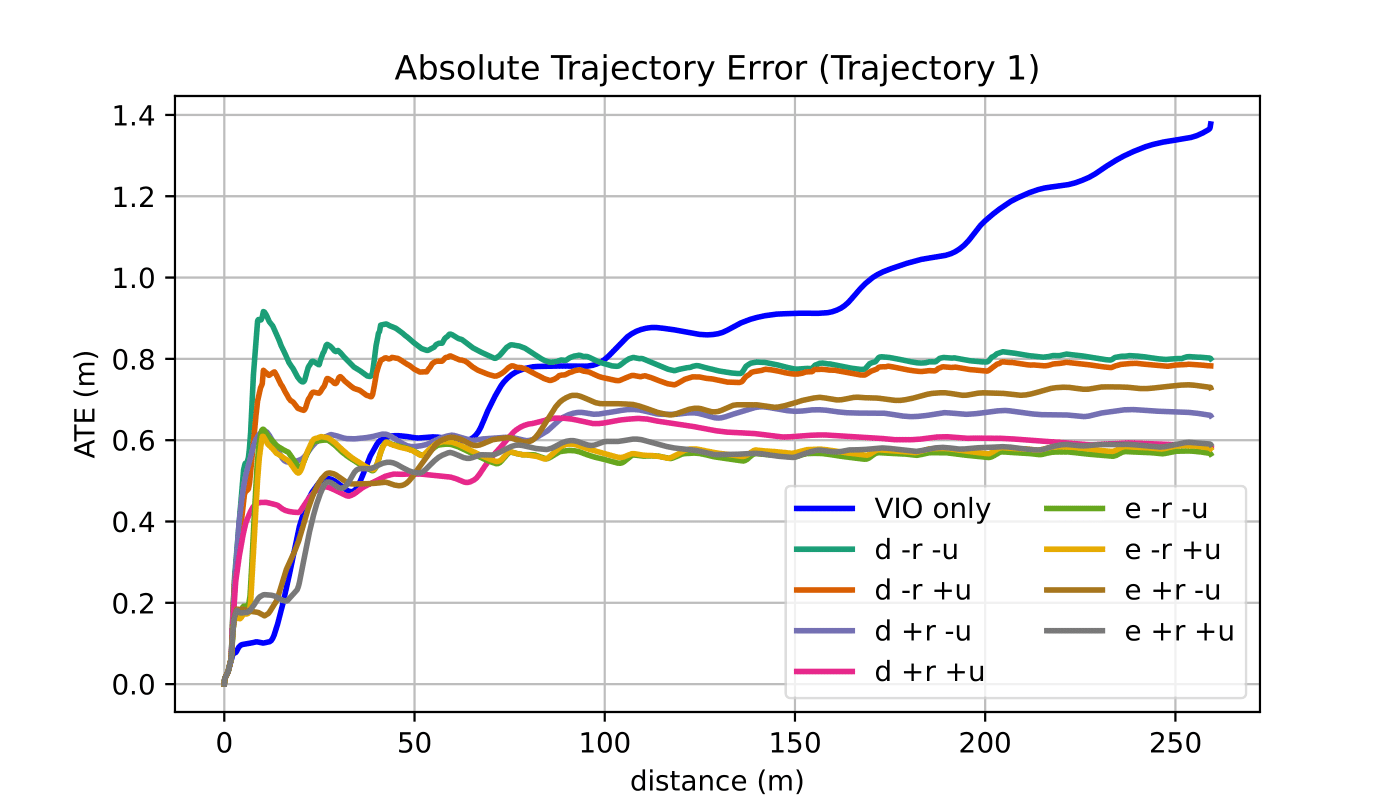}
	\caption{Absolute Trajectory Error (ATE) for VIO with and without camera pose regressors over the travel distance of 'Trajectory 1'. `d' (resp. 'e') denotes `dropout' (resp. `ensemble'), `+u' (resp. `+r') indicates uncertainty prediction (resp. outlier rejection) is applied, and `-' signifies the absence of the respective feature.} \label{fig:ATE}
	\vspace{-0.2in}
\end{figure}

\begin{table*}[t]
	\caption{Average positional and rotational errors for different VIO configurations over all test trajectories (m/degrees). The blue and red bold numbers indicate the first and second-best performances, respectively, for each test case, excluding the `only VIO' configuration.}
	\label{tab:VIO_accuracy}
	\centering
	\begin{tabular}{c || c || c | c || c | c || c | c || c | c }
		\toprule
		\multirow{ 2}{*}{\textbf{Trajectory}} & \multirow{ 2}{*}{\textbf{only VIO}} & 
		\multicolumn{2}{c||}{\textbf{constant + no rejection}} & \multicolumn{2}{c||}{\textbf{estimated + no rejection}} &
		\multicolumn{2}{c||}{\textbf{constant + rejection}} & \multicolumn{2}{c}{\textbf{estimated + rejection}} \\ \cline{3-10}

		{} & {} & \textbf{dropout} & \textbf{ensemble} & \textbf{dropout} & \textbf{ensemble} & \textbf{dropout} & \textbf{ensemble} & \textbf{dropout} & \textbf{ensemble} \\
		\midrule
		\textbf{1} & 1.224/1.01 &
		0.659/5.76 & \second{0.468}/2.43 &
		0.659/6.14 & 0.472/2.88 &
		0.942/1.44 & 0.601/\second{1.40} &
		0.470/\first{1.35} & \first{0.418}/1.51 \\
		\textbf{2} & 4.104/3.13 &
		1.225/6.80 & 0.771/5.28 &
		1.205/12.43 & 0.762/4.53 &
		0.942/\first{1.86} & \second{0.601}/2.92 &
		0.882/\second{1.92} & \first{0.524}/1.99 \\
		\textbf{3} & 11.917/4.58 &
		1.201/6.16 & 0.750/4.22 &
		1.241/6.40 & 0.724/3.61 &
		0.947/1.98 & \second{0.601}/2.24 &
		0.760/\first{1.72} & \first{0.510}/\second{1.97} \\
		\textbf{4} & 1.411/1.80 &
		1.221/8.88 & 0.793/5.19 &
		1.069/7.70 & 0.746/5.26 &
		0.999/5.02 & \first{0.678}/\second{2.39} &
		\second{0.720}/\first{1.75} & 0.815/5.39 \\
		\textbf{5} & 6.476/2.04 &
		0.918/7.17 & 0.543/3.13 &
		0.802/10.73 & 0.548/3.79 &
		0.846/2.55 & \second{0.484}/\second{1.91} &
		0.512/2.05 & \first{0.455}/\first{1.90} \\
		\textbf{6} & 6.876/2.12 &
		0.902/7.37 & \second{0.586}/4.43 &
		0.806/6.21 & \first{0.557}/4.47 &
		0.950/\second{3.23} & 0.660/\first{2.98} &
		0.686/4.50 & 0.586/4.60 \\
		\textbf{7} & 1.767/1.35 &
		0.836/4.56 & 0.567/3.29 &
		0.715/5.30 & \second{0.513}/3.35 &
		0.786/1.95 & \first{0.426}/1.97 &
		0.724/2.10 & 0.553/2.42 \\ \hline
		\textbf{Average} & 4.824/2.29 &
		0.995/6.67 & 0.640/3.99 &
		0.928/7.84 & 0.617/3.98 &
	 	0.869/2.58 & \second{0.592}/\second{2.26} &
	 	0.679/\first{2.20} & \first{0.552}/2.82  \\
		\bottomrule
	\end{tabular}
\vspace{-0.1in}
\end{table*}

\textbf{Results.}
Table \ref{tab:VIO_accuracy} presents the mean positional and rotational errors for each evaluated scenario across all trajectories. Figure \ref{fig:traj} illustrates the actual and estimated trajectories for the 'only VIO' and 'ensemble + estimated + rejection' scenarios, specifically for Trajectories 1, 3, 5, and 7. Figure \ref{fig:ATE} displays the Absolute Trajectory Errors (ATE) for all scenarios along 'Trajectory 1'.

Fig. \ref{fig:traj} and \ref{fig:ATE} demonstrate that a drift accumulates in the 'only VIO' scenario, resulting in an increase in positional error. On the other hand, when the VIO is integrated with the camera pose regressors, errors tend to converge, as the camera pose regressors effectively constrain the error bounds. The effectiveness of the pose regressors is evident in Table \ref{tab:VIO_accuracy}, where the positional error for the 'only VIO' case is significantly higher than in other cases.

Table \ref{tab:VIO_accuracy} represent that `ensemble', `estimated', and `rejection' exhibit better positional accuracy compared  `dropout', `constant', and `no rejection', respectively. Specifically, the `ensemble + estimated + rejection' scenario achieves the lowest average positional error , showing a 44.5\% improvement in accuracy over the `dropout + constant + no rejection' scenario which displays the highest positional error. Fig. \ref{fig:ATE} further illustrates that `ensemble + estimated + rejection' configuration not only converges the lowest error along with `dropout + estimated + rejection', `ensemble + constant + no rejection', and  `ensemble + estimated + no rejection' configurations but also maintains a lower error than the configurations prior to convergence. As a result, the uncertainty quantification of the camera pose regressor and its integration into pose graph optimization, both as a covariance measure or through outlier rejection, enhances the overall positional performance.

Rotational errors are lower when `rejection' configurations are applied, compared to `no rejection', which is noteworthy as it suggests that position-based rejection also enhances rotational performance. This is because there exists a correlation between positional and rotational uncertainties, as indicated in \cite{Kendall2016ICRA}. Furthermore, `rejection' scenarios exhibit rotational accuracies that are relatively comparable to the `Only VIO' scenario, while other configurations tend to underperform relative to `Only VIO' with respect to rotation. However, `estimated' scenarios do not consistently outperform the `constant' scenarios in terms of rotational accuracy.

In summary, the experiments demonstrate that integrating the camera pose regressor into the VIO framework prevents drift. Moreover, applying covariance and outlier rejections, derived from uncertainty quantification, improves the positional accuracy. Additionally, outlier rejection safeguards against the potential decline in rotational accuracy that may result from using the pose regressor.

\section{Conclusions}

In this research,  we introduced a framework that capitalizes on the capabilities of Neural Radiance Fields (NeRF) to enhance real-time and robust localization for robotic navigation by integrating a camera pose regressor, trained on augmented data generated from NeRF, with a VIO framework. The uncertainty of the pose regressor was quantified in a Bayesian manner, estimating the potential error between the real and estimated poses caused by discrepancies between the actual scene and its NeRF representation. This quantification was integrated into the VIO framework, enhancing its reliability and robustness. Our experimental validation with Flightgoggle simulation confirms that incorporating NeRF-aided camera pose regression, alongside uncertainty quantification, substantially improves the VIO framework's performance. Future research will explore the scalability of our method to larger environments, such as entire indoor floors, and investigate effectiveness of uncertainty quantification given the environmental changes over time after the NeRF training.







\bibliographystyle{unsrt}
\bibliography{reference}

\end{document}